\documentclass{article}

\pdfoutput=1 

\usepackage{arxiv}

\usepackage[utf8]{inputenc} 
\usepackage[T1]{fontenc}    
\usepackage{hyperref}       
\usepackage{url}            
\usepackage{booktabs}       
\usepackage{amsfonts}       
\usepackage{nicefrac}       
\usepackage{microtype}      
\usepackage{graphicx}

\usepackage{authblk}

\usepackage{caption}
\usepackage{subcaption}

{

\title{Detection and Annotation of Plant Organs from Digitized Herbarium Scans using Deep Learning}

\author[1,2]{\textbf{Sohaib~Younis} \thanks{Corresponding author: Sohaib Younis, sohaibyounis89@gmail.com}}
\author[1,3]{\textbf{Marco~Schmidt}}
\author[1]{\textbf{Claus~Weiland}}
\author[4]{\textbf{Stefan~Dressler}}
\author[2]{\textbf{Bernhard~Seeger}}
\author[1]{\textbf{Thomas~Hickler}}
\affil[1]{Senckenberg Biodiversity and Climate Research Centre (SBiK-F), Frankfurt am Main, Germany}
\affil[2]{Department of Mathematics and Computer Science, Philipps-University Marburg, Marburg, Germany}
\affil[3]{Palmengarten der Stadt Frankfurt, Frankfurt am Main, Germany}
\affil[4]{Senckenberg Research Institute and Natural History Museum, Frankfurt am Main, Germany}

\date{}

\renewcommand{\headeright}{}
\renewcommand{\undertitle}{}

\hypersetup{
pdftitle={Detection and Annotation of Plant Organs from Digitized Herbarium Scans using Deep Learning},
pdfauthor={Sohaib~Younis, Marco~Schmidt, Claus~Weiland, Stefan~Dressler, Bernhard~Seeger, Thomas~Hickler},
pdfkeywords={Herbarium Specimens, Plant Organ Detection, Deep Learning, Convolutional Neural Networks, Object Detection and Localization, Image Annotation, Digitization},
}

\renewcommand\footnotemark{}

\begin{document}
\maketitle

\begin{abstract}
	As herbarium specimens are increasingly becoming digitized and accessible in online repositories, advanced computer vision techniques are being used to extract information from them. The presence of certain plant organs on herbarium sheets is useful information in various scientific contexts and automatic recognition of these organs will help mobilize such information. In our study we use deep learning to detect plant organs on digitized herbarium specimens with Faster R-CNN. For our experiment we manually annotated hundreds of herbarium scans with thousands of bounding boxes for six types of plant organs and used them for training and evaluating the plant organ detection model. The model worked particularly well on leaves and stems, while flowers were also present in large numbers in the sheets, but not equally well recognized.
\end{abstract}


\section{Introduction}
Herbarium collections have been the base of systematic botany for centuries. More than 3000 herbaria are active on a global level, comprising c. 400 Mio specimens, a number that doubled since the early 1970s and is growing steadily \cite{thiers2020world}. Accessibility of these collections has been improved by international science infrastructure aggregating specimen data and increasingly also digital images of the specimens. Plant specimens, being usually flat and of a standard format approximating A3 size, are easier to digitize than most other biological collection objects. The Global Plants Initiative \cite{smith2014global} has been very successful in digitizing type specimens around the world, single collections like the National Museum of Natural History in Paris have digitized their collections completely \cite{le2017french} and large scale national or regional digitization initiatives are already taking place or are planned for the near future \cite{borsch2020complete}. Presently there are more than 27 million plant specimen records with images available via the GBIF platform (www.gbif.org), the vast majority of these images being herbarium scans.

This rising number of digitized herbarium sheets provides an opportunity to employ computer-based image processing techniques like deep learning to automatically identify species and higher taxa \cite{carranza2017going, younis2018taxon, carranza2018automated} or to extract other useful information from the images, like the presence of pathogens (as done for live plant photos by Mohanty et al. 2016 \cite{mohanty2016using}). Deep learning is a subset of machine learning methods for learning data representation. Deep learning techniques require huge amounts of training data to learn the features and representation of that data for the specified task by fine tuning parameters of hundreds or thousands of neural networks, arranged in multiple layers. Learning the value of these parameters can take vast computer and time resources, especially on huge datasets.

The most common type of deep learning network architecture being used for extracting image features is the Convolutional Neural Network (CNN) \cite{lecun1995convolutional}. The layers and connectivity of neurons is inspired by the biological process of the animal visual cortex \cite{matsugu2003subject, hubel1968receptive}. A convolutional neural network extracts the features of an image by passing through a series of convolutional, nonlinear, pooling (image downsampling) layers and passes them to a fully connected layer to get the desired output. Each convolutional layer extracts the visual features of the image by applying convolution operations to the image with kernels, using a local receptive field, to produce feature maps and passing it as input to the next layer. The initial layers in the network compute primitive features on the image such as corners and edges, the deeper layers use these features to compute more complex features consisting of curves and basic shapes and the deepest layers combine these shapes and curves to create recognizable shapes of objects in the image \cite{yosinski2014transferable, zeiler2014visualizing}.

In this paper we use deep learning for detecting plant organs on herbarium scans. The plant organs are detected using an object detection network, which works by localizing each object with a bounding box on the image and classifying it. There are many types of networks based on CNN, used for this application. In this study, a network called Faster R-CNN \cite{ren2015faster} was used, which is part of the R-CNN family for object detection. Region-based Convolutional Networks (R-CNN) identify objects and their locations in an image. Faster R-CNN networks have shown state of the art performance in various object detection applications and competitions \cite{zhao2019object}. Therefore many researchers have explored the use of Faster R-CNN for detecting various plant organs like flowers, fruits and seedlings \cite{sa2016deepfruits, stein2016image, hani2020comparative, mai2018faster, sun2018detection, bargoti2017deep, jiang2019deepseedling}. To our knowledge this is the first time object detection has been used for detecting multiple types of plant organs on herbarium scans.  Identifying and localizing plant organs on herbarium sheets is a first necessary step for some interesting applications. The presence and state of organs like flowers and fruits can be used in phenological studies over long time periods and may give us more insight into climate change effects since the time of the industrial revolution \cite{willis2017old, lang2019using}.

\section{Methods}

\subsection{Network architecture}

A typical object detection network consists of object localization and classification integrated into one convolutional network. There are two main types of meta-architectures available for this application: single stage detectors like Single Shot Multibox Detectors (SSD) \cite{liu2016ssd} and You only look once (YOLO) \cite{redmon2016you}, and two-stage, region-based CNN detectors like Faster R-CNN. Single stage detectors use a single feed- forward network to predict object class probabilities along with bounding box coordinates on image. Faster R-CNN is composed of three modules: 1) a deep CNN image feature extraction network, 2) a Region Proposal Network (RPN), used for detection of a predefined number of Regions of Interests (RoIs) where the object(s) of interest could reside within the image, followed by 3) Fast R-CNN \cite{girshick2015fast}, computes a classification score along with class-specific bounding box regression for each of these regions.

\begin{figure}[h!]
\centering
      \includegraphics[width=0.6\textwidth]{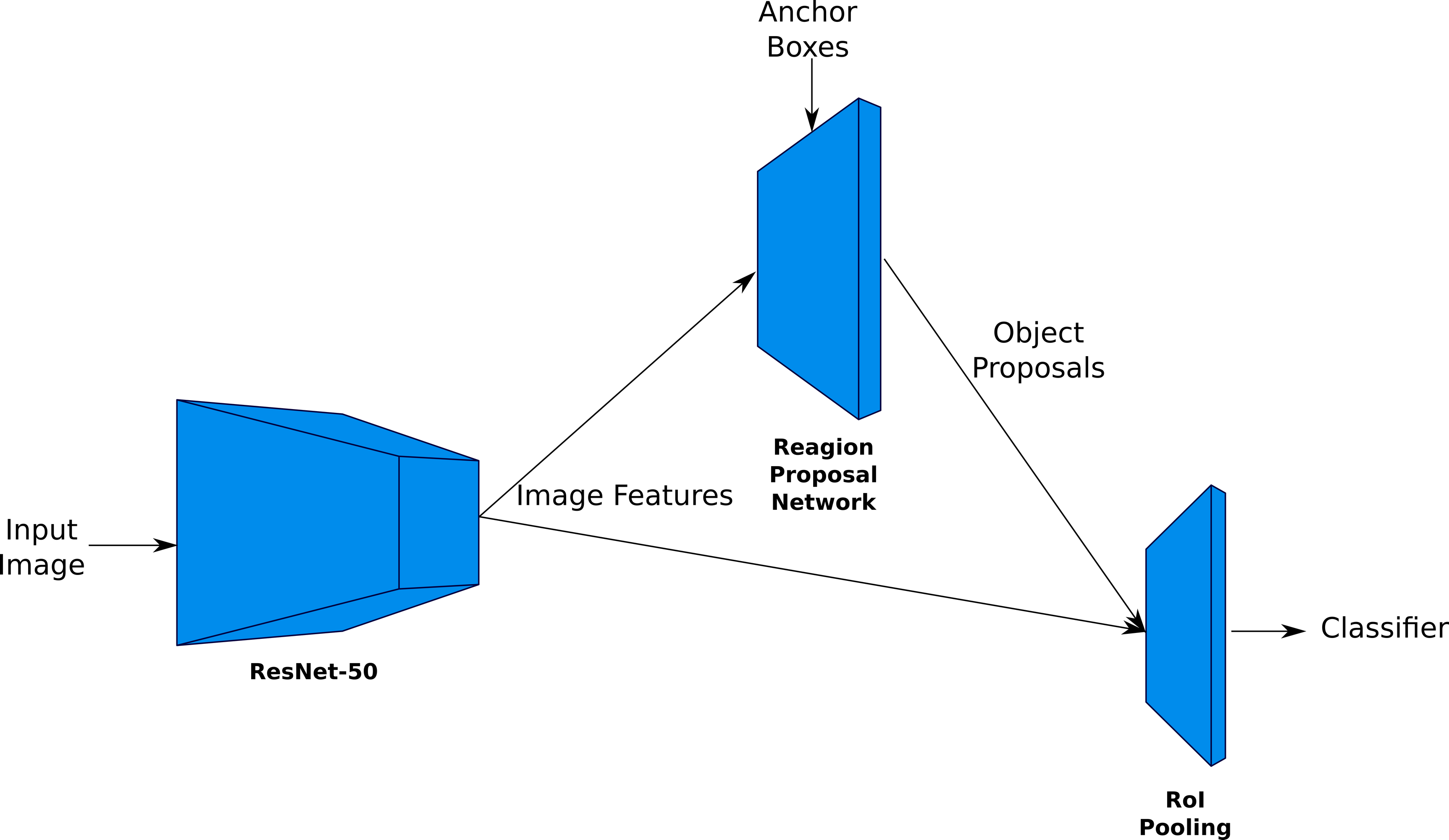}    
  	\caption{An illustration of the Faster R-CNN architecture, with ResNet for image feature extraction, RPN for generating object proposals and RoI Pooling for creating fixed-size feature maps for each proposal.}
	\label{figure1}
\end{figure}

The CNN feature extraction network used in this paper is based on the ResNet-50 architecture \cite{he2016deep}, without the final fully connected layer. The Region Proposal Network creates thousands of prior or anchor boxes to estimate the location of objects in the image. The anchor boxes are predefined bounding boxes of certain height and width tiled across the image, determined by their scale and aspect ratios, in order to capture different sizes of objects of specific classes. The RPN generates these proposals by adjusting these anchors with coordinate offsets of the object bounding boxes and predicts the possibility of each anchor being a foreground object or a background. These proposals are sorted according to their score and top N proposals are selected by Non-Maximum Suppression (NMS), which are then passed to Fast R-CNN stage. NMS reduces the high number of proposals for the next stage by short listing the proposals with highest score having minimum overlap with each other by removing the proposals with overlap above a predefined threshold for each category. In the next stage the proposals with feature maps of different shapes are pooled with a ROI pooling layer, which performs max-pooling on the inputs of non-uniform sizes to obtain a fixed number of uniform size feature maps. These feature maps are propagated through fully connected layers, which end in two siblings fully connected layers for object classification and bounding box regression respectively. An illustration of Faster R-CNN is shown in Figure \ref{figure1}.

\subsection{Image Annotation}

\begin{figure}[h]
\centering
      \includegraphics[width=0.5\textwidth]{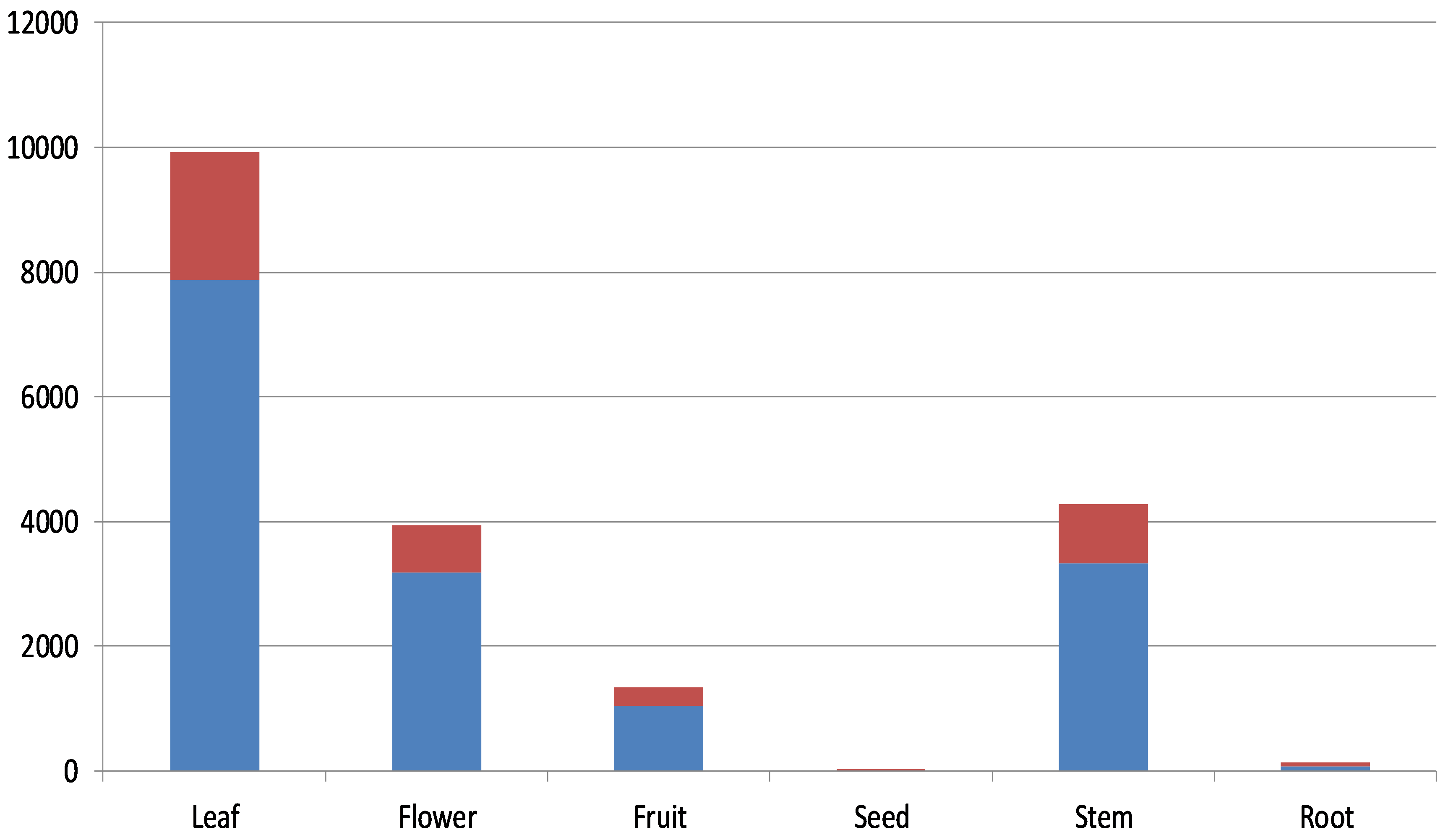}    
  	\caption{A column chart showing the number of annotated bounding boxes for each organ. Red: Test subset, Blue: Training subset}
	\label{figure2}
\end{figure}

The herbarium scans annotated for training the object detection network were selected from the MNHN (Muséum national d’Histoire naturelle) vascular plant herbarium collection dataset in Paris \cite{le2017french}, from open access images contributed to the GBIF portal. A total of 653 images were downloaded and rescaled from their original average size of c. 5100 by 3500 pixels to 1200 by 800 pixels, in order to preserve the aspect ratio of the scans and to speed up the learning by reducing the number of pixels. All these images were annotated for six different types of organs using LabelImg \cite{tzutalin2015}, a Python graphical toolkit for image annotation using bounding boxes. The average rate for manual image annotation was 8 to 15 herbarium sheets per hour, depending on the difficulty and number of bounding boxes to be annotated. The total number of annotated bounding boxes for all 653 images was 19654, with an average of 30.1 bounding boxes per image. From these 653 annotated images, 155 of them were either annotated or verified by an expert, making a validated subset hence used for testing and the 498 were used for training, as shown in Figure \ref{figure2} and in more detail in Table \ref{table1}.

\begin{table}[h]	
	\centering
	\begin{tabular}{ |c|c|c|c| }
		\hline
		\textbf{Category} & \textbf{Training subset} & \textbf{Test subset} & \textbf{Complete dataset} \\
					& (498 images) & (155 images) & (653 images) \\
		\hline
		{Leaf} & 7886 & 2051 & 9937 \\
		{Flower} & 3179 & 763 & 3942 \\
		{Fruit} & 1047 & 296 & 1343 \\
		{Seed} & 4 & 6 & 10 \\
		{Stem} & 3323 & 961 & 4284 \\
		{Root} & 78 & 60 & 138 \\
		\hline
		\textbf{Total} & 15517 & 4137 & 19654 \\
		\hline
	\end{tabular}
	\caption{The number of annotated bonding boxes for each plant organ in training and test subset.}
	\label{table1}
\end{table}

Preparing our data was not always straight-forward. The manual localization and labelling of plant organs from specimens encountered the following difficulties:  Buds, flowers and fruits are different stages emerging in the life cycle of plant reproductive organs and in some cases it was therefore difficult to find a clear distinction between these structures.  In some taxa, different plant organs were impossible to separate being small and crowded, e.g. in dense inflorescences with bracts and flowers, or stems densely covered by leaves. In a few cases it was also hard to differentiate from the digital image between roots and stolons or other stem structures. In all these cases we placed our labelled boxes in a way to best characterize the respective plant organ. Sometimes this involved including parts of other organs, sometimes, if sufficient clearly assignable material was available, difficult parts were left out.

\subsection{Implementation}

\begin{figure}[h]
\centering
      \includegraphics[width=0.5\textwidth]{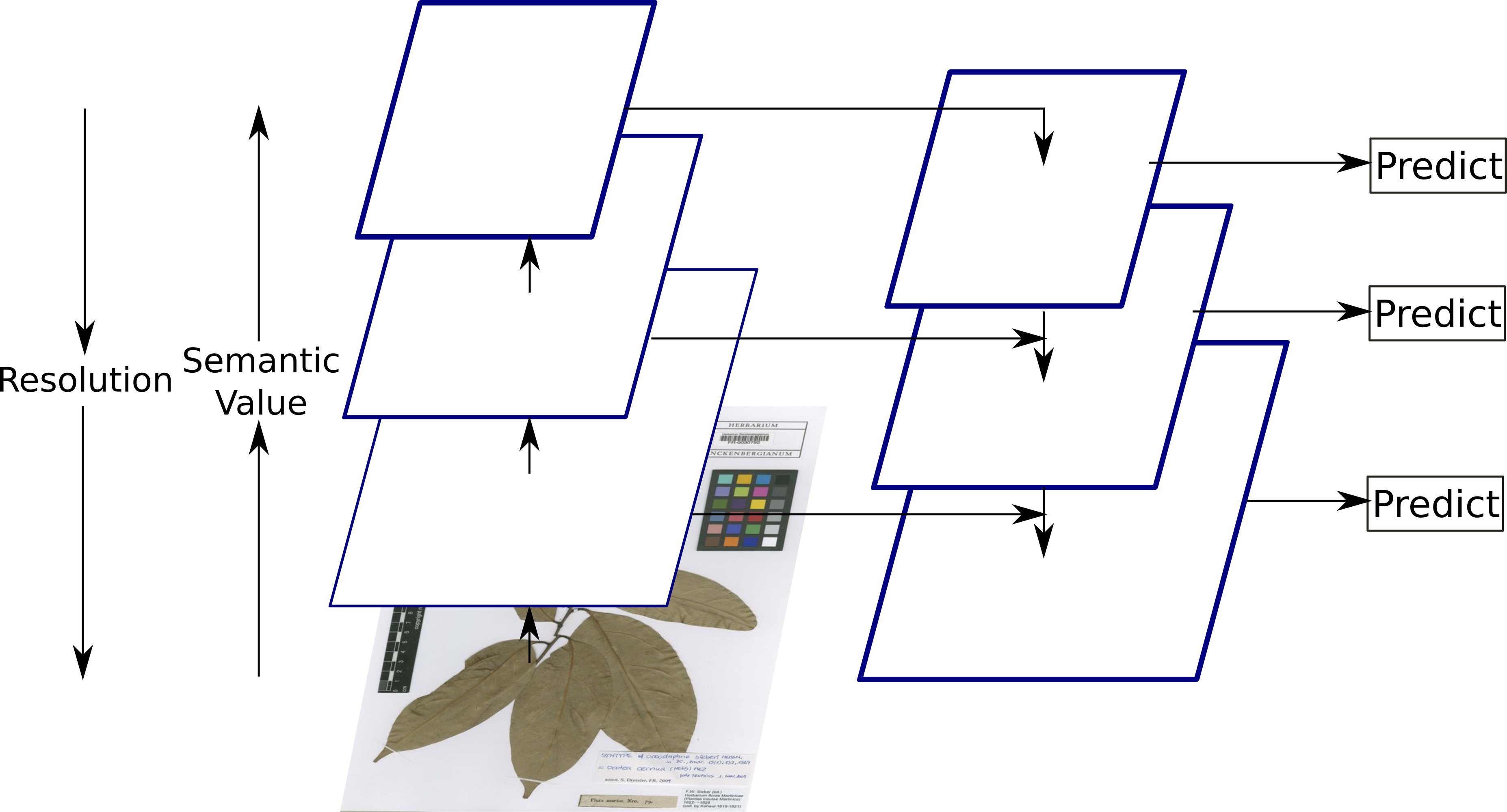}    
  	\caption{An illustration of Feature Pyramid Network, where feature maps are indicated by blue outlines and thicker outlines denote semantically stronger features \cite{lin2017feature}.}
	\label{figure3}
\end{figure}

The object recognition task was performed using Faster R-CNN, as described in the network architecture, with Feature Pyramid Network \cite{lin2017feature} backbone. Feature Pyramid Network increases the accuracy of the object detection task by generating multi-scale feature maps from a single scale feature map of ResNet output, by making top-down pathways in addition to the usual bottom-up pathways used by a regular convolutional network for feature extraction, where each layer of the network represents one pyramid level. The bottom–up pathway increases the semantic value of the image features, from corners and edges in the initial layers to detecting high level structures and shapes of objects in the image in the final layers, while reducing its resolution at each layer. The top-down pathway then reconstructs higher resolution layers from the most semantically rich layer, with predictions made independently at all levels as shown in Figure \ref{figure3}. This approach provides Faster R-CNN with feature maps at different resolutions for detecting objects of multiple scales.

In order to reduce the training time and more importantly because of the small size of the training dataset, transfer learning \cite{yosinski2014transferable} was implemented to initialize the model weights pre-trained on ImageNet dataset \cite{deng2009imagenet}. Since the initial layers of a CNN usually learn very generic features that can also be used in new contexts, pre-trained weights can initialize the weights for these layers. For the deeper layers transfer learning is used to initialize the parameter weights pre-trained on the ImageNet dataset and then fine-tuned during training using the annotated herbarium scan dataset till convergence.

The model was implemented with the Detectron2 \cite{wu2019detectron} library in PyTorch framework and trained using Stochastic Gradient Descent optimizer with a learning rate of 0.0025 and momentum of 0.9. The anchor generator in the Region Proposal Network (see section above on network architecture) had 6 anchor scales [32, 64, 128, 256, 512, 1024] (square root of area in absolute pixels) each with 3 aspect ratios of [1:2, 1:1, 2:1]. The thresholds for non-maximum suppression (NMS) were 0.6 for training and 0.25 for testing respectively.

Because of the large image size and additional parameters of Faster R-CNN, a minibatch size of 4 images per GPU (TITAN Xp) was selected for training the model. The model was trained twice, once with a training subset of 498 images on a single GPU for 9000 iterations and performance evaluated on the test subset of 155 images, also on a single GPU. The model was then trained again on all 653 annotated images on 3 GPUs for 18000 iterations for predicting plant organs on another un-annotated dataset. This dataset consists of 708 full scale herbarium scans, with an average size of c. 9600 by 6500 pixels, from the Herbarium Senckenbergianum (HS) \cite{otte2011herb}.

\section{Results}

The minimum threshold for any prediction to be acceptable was having a score (probability) of 0.5. The performance of the model was evaluated with Average Precision metric using the Pascal VOC 2012 \cite{everingham2011} and COCO methods \cite{lin2014microsoft}. The performance of the model on the test subset, trained on MNHN training subset, are shown in Table \ref{table2}. The Pascal VOC method considers all predictions as positive that have Intersection over Union of at least 0.5. Intersection over Union (IoU) is a measure to calculate the overlap of the predicted bounding boxes with the ground truth bounding boxes. If multiple detections of the same object are detected, it counts the first one as a positive while the rest as negatives. The Average Precision is calculated by estimating the area under the precision-recall curve for all correct predictions, giving a score between 0 and 1. This metric for Average Precision with IoU of 0.5 is called AP50. In the COCO method, AP is calculated with three metrics all having values between 0 and 100. The first metric is the same as Pascal VOC called AP50. The second metric is AP75, with a minimum IoU of 0.75, and the third is AP is the average over 10 IoU levels from 0.5 to 0.95 with a step size of 0.05. This method also gives the AP for each category, as shown in Table \ref{table3}, along with the total bounding boxes for each category in the test subset. The difference between the Pascal VOC and COCO for AP50 is most likely due to different methods to calculate the Average Precision.

\begin{table}[h!]	
	\centering
	\begin{tabular}{ |c|c|c|c| }
		\hline
		{AP50 (Pascal VOC)} & {AP50 (COCO)} & {AP75} & {AP} \\
		\hline
		0.54 & 22.8 & 6.8 & 9.7 \\
		\hline
	\end{tabular}	
	\caption{Results of the model evaluation on the test subset with Pascal VOC and COCO average precision methods.}
	\label{table2}
\end{table}

\begin{table}[h!]	
	\centering
	\begin{tabular}{ |c|c|c| }
		\hline
		{Category} & {Bounding Boxes} & {AP} \\
		\hline
		{Leaf} & 2051 & 26.5 \\
		{Flower} & 763 & 4.7 \\
		{Fruit} & 296 & 7.8 \\
		{Seed} & 6 & 0.0 \\
		{Stem} & 961 & 9.9 \\
		{Root} & 60 & 9.4 \\
		\hline
	\end{tabular}	
	\caption{The number of annotated bonding boxes for each plant organ in training and test subset.}
	\label{table3}
\end{table}

\begin{figure}[h!]
\begin{subfigure}{.48\textwidth}
  \centering
  \includegraphics[width=.8\linewidth]{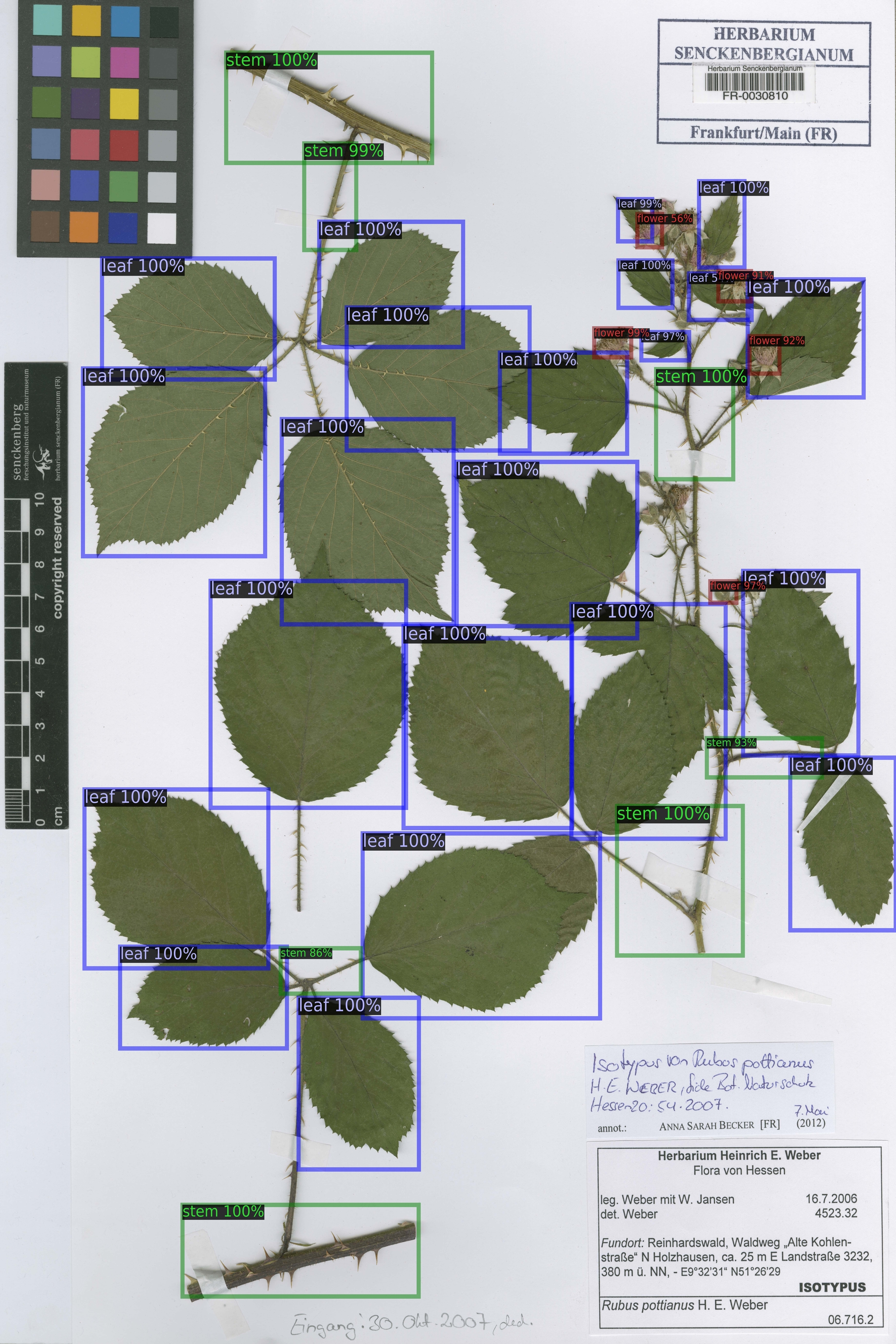}  
\end{subfigure}
\begin{subfigure}{.48\textwidth}
  \centering
  \includegraphics[width=.8\linewidth]{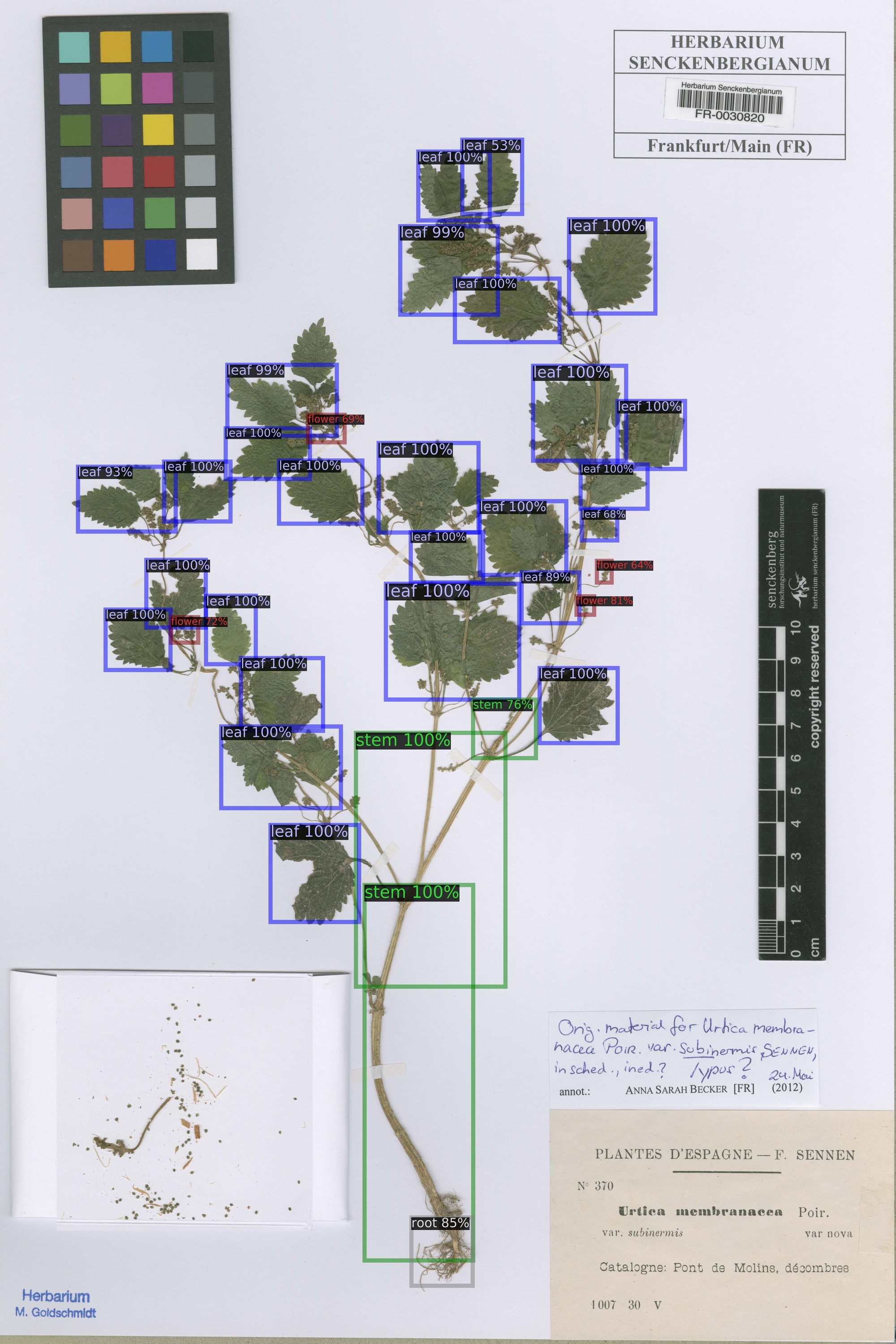}  
\end{subfigure}


\begin{subfigure}{.48\textwidth}
  \centering
  \includegraphics[width=.8\linewidth]{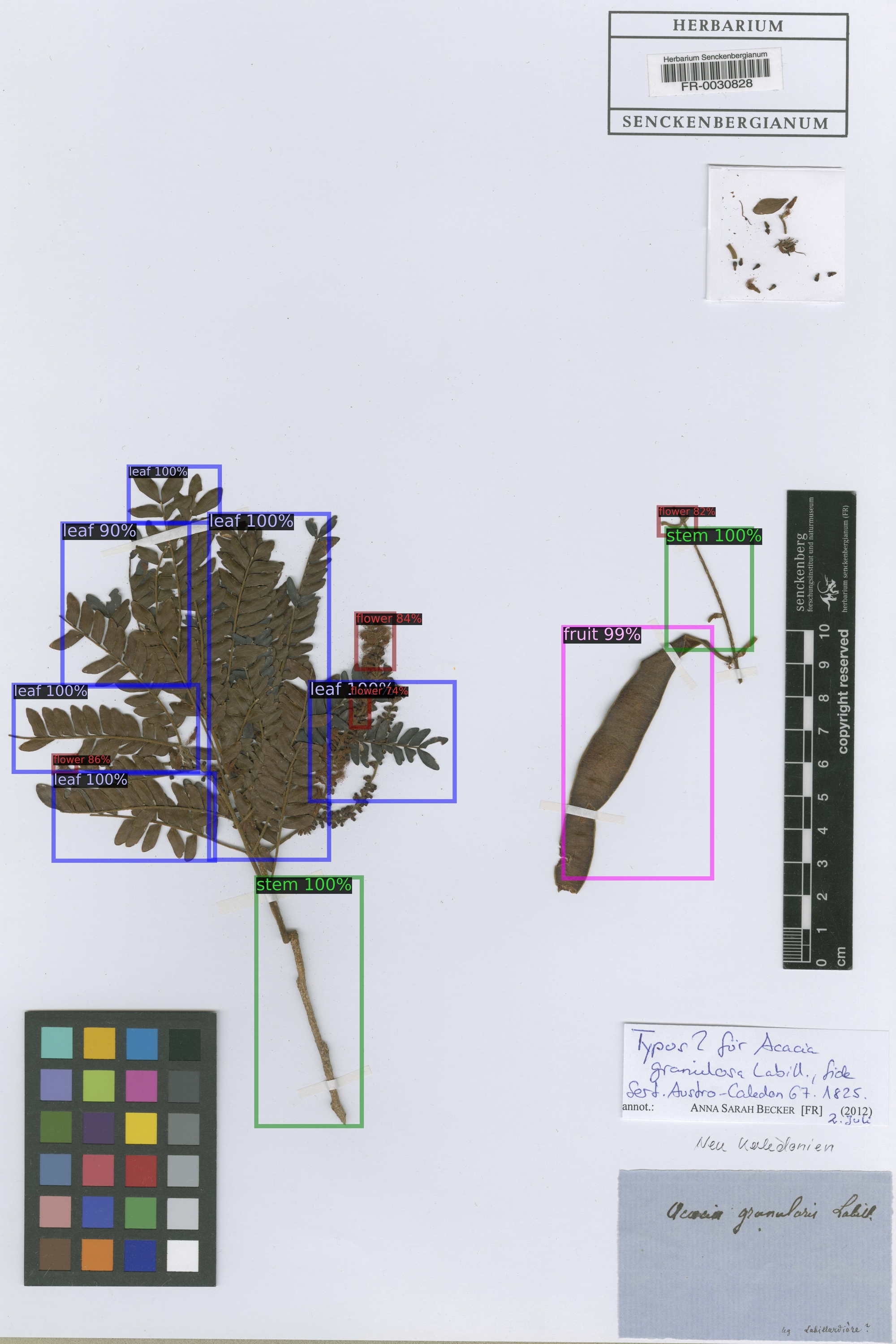}  
\end{subfigure}
\begin{subfigure}{.48\textwidth}
  \centering
  \includegraphics[width=.8\linewidth]{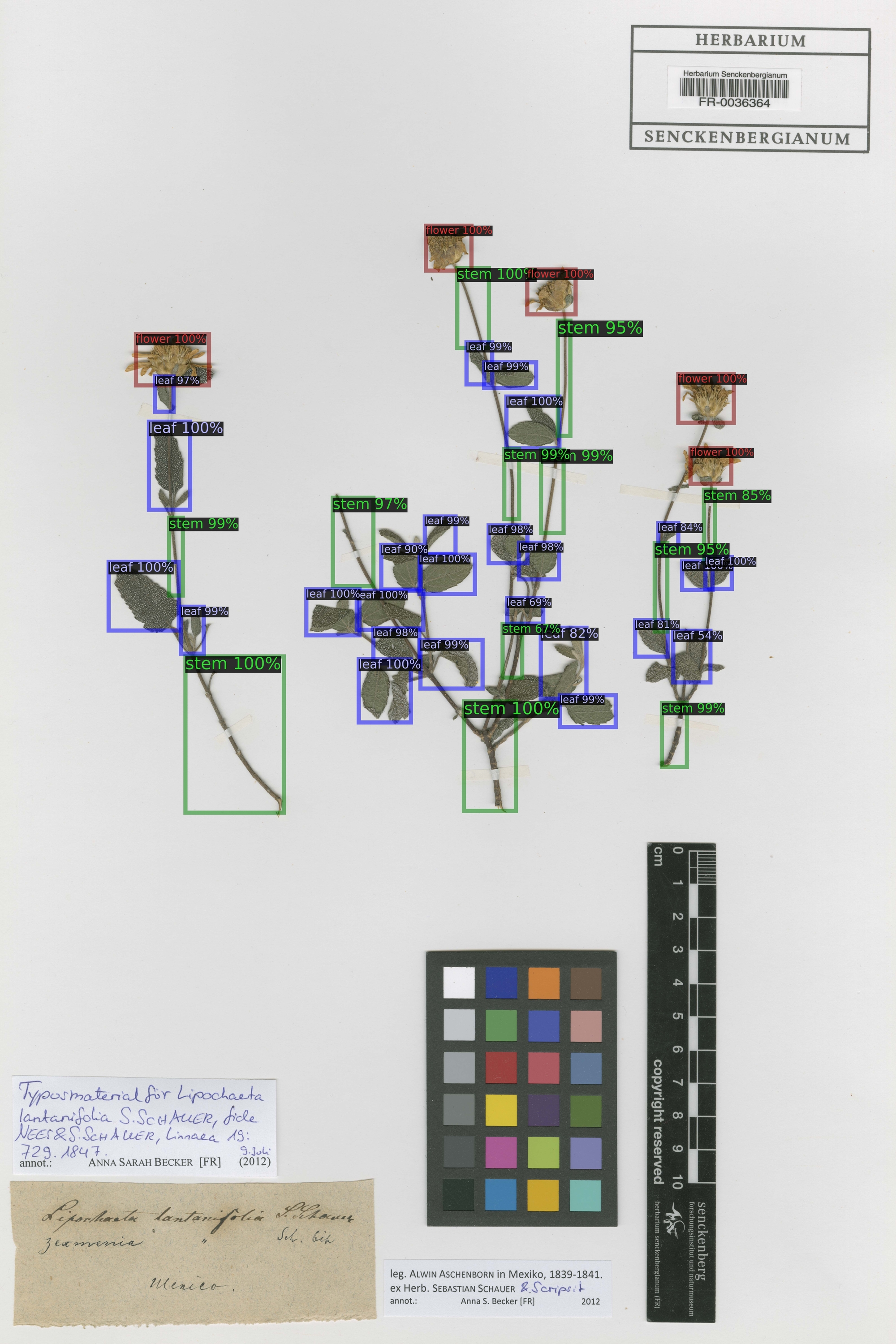}  
\end{subfigure}
\caption{Sample results of organ detection performed on unseen full scale Herbarium Senckenbergianum scans. Color scheme for bounding boxes is; Leaf:Blue, Flower:Maroon, Fruit:Magenta, Seed:Yellow, Stem:Green, Root:Gray}
\label{figure4}
\end{figure}

A sample result of the model on the HS dataset, trained on 653 annotated MNHN images is shown in Figure \ref{figure4}. The organ detection model was sucessfully able to detect almost all of plant organs in majority of images. Out of these 708 scans, 203 were annotated based on the predictions of the organ detection to test the model performance. The dataset of these 203 herbarium scans, along with the result of detections and the annotations are avaialble at PANGAEA (\href{https://doi.pangaea.de/10.1594/PANGAEA.920895}{\emph{link}}) \cite{younis2020poda}.

\begin{table}[h!]
	
	\centering
	\begin{tabular}{ |c|c|c| }
		\hline
		{AP50} & {AP75} & {AP} \\
		\hline
		31.9 & 25.7 & 26.4 \\
		\hline
	\end{tabular}
	
	\caption{Result of model evaluation on Herbarium Senckenbergianum annotated dataset.}
	\label{table4}
\end{table}

\begin{table}[h!]
	
	\centering
	\begin{tabular}{ |c|c|c| }
		\hline
		{Category} & {Bounding Boxes} & {AP} \\
		\hline
		{Leaf} & 3330 & 52.9 \\
		{Flower} & 1899 & 29.9 \\
		{Fruit} & 153 & 11.3 \\
		{Seed} & 2 & 0.0 \\
		{Stem} & 1055 & 53.0 \\
		{Root} & 77 & 11.0\\
		\hline
	\end{tabular}
	
	\caption{Average Precision (AP COCO) of each type of organ along with the total bounding boxes for each category in the Herbarium Senckenbergianum annotated dataset.}
	\label{table5}
\end{table}

The performance of the model on the annotated Herbarium Senckenbergianum dataset is shown in Table \ref{table4} and Table \ref{table5}. The average precision on these 203 scans is generally higher than the MNHN test subset, there are two main reason for this 1) The organ detection model for full scale detection was trained on all 653 images of MNHN annotated dataset before detection on HS dataset, 2) The annotation of these 203 images from HS dataset were done based on the predictions of organs on scans as shown in Figure \ref{figure4}.

\section{Discussion}

This paper presents a method to detect multiple types of plant organs on herbarium scans. For this research we annotated hundreds of images with thousands of bounding boxes with hand for each possible plant organ. A subset of these annotated scans was then used for training a deep learning for organ detection. After training the model was used to predict the type and location of plant organs on the test subset. The automated detection of plant organs in our study was most successful for leaves and stems (Table \ref{table3} and Table \ref{table5}). Best AP values for leaves are likely due to the largest set of annotated bounding boxes. Good values for stems and roots may be explained by the relative uniformity of these organs throughout the plant kingdom, as compared to the morphologically more diverse flowers, and fruits in between these. Seeds are rarely visible on herbarium sheets and require more training material.

The model was trained again on all the annotated scans earlier and tested on a different un-annotated dataset. The model performed well based on visual inspection. In order to evaluate the performance of the model with average precision metric, around 200 of these scans were annotated by hand based on the predicted bounding boxes. The predicted bounding boxes dramatically reduced the time to annotate these scans, since the predictions for leaves and stems were fairly accurate. After being annotated these scans were compared with the predictions to evaluate the precision of the organ detection model on this dataset.

Most computer vision approaches on plants focus on live plants, often in the context of agriculture or plant breeding and therefore including only a limited set of taxa. The present approach not only targets a much larger group of organisms and morphological diversity, comparable to applications in citizen science \cite{waldchen2019flora}, but can also be applied on a wider time scale by including collection objects from hundreds of years of botanical research. Some significant recent similar approaches to detect plant organs on herbarium scans are GinJinn \cite{ott2020ginjinn} and LeafMachine \cite{weaver2020leafmachine}. GinJinn uses an object-detection pipeline for automated feature extraction from herbarium specimens. This pipeline can be used to detect any type of plant organ, which the authors of this research demonstrated by detecting leaves on a sample dataset. LeafMachine is another approach, which tries to automate extraction of leaf traits, like class, size and number, from digitized herbarium specimens with machine learning.

\section{Conclusions}

Our present work is focussing on the detection of plant organs from specimen images. The presence of flowers and fruits on specimens is a new source of data for phenological studies \cite{willis2017old} interesting in the context of climate change. Presence of roots would identify plant specimens potentially containing root symbionts like mycorrhizal fungi or N-fixing bacteria for further study by microbiological or genetic methods \cite{heberling2019utilizing}. Up to now, this requires visual examination of the specimens by humans, an automated approach using computer vision would considerably reduce the effort. Furthermore, the detection and localization of specific plant organs on a herbarium sheet would also enable or improve further computer-vision applications, including quantitative approaches based on counting these organs, improved recognition of qualitative organ-specific traits like leaf shape as well as quantitative measures such as leaf area or fruit size.

Localization of plant organs will improve automated recognition and measurements of organ-specific traits, by preselecting appropriate training material for these approaches. The general approach of measuring traits from images instead of the specimen itself, has been shown to be precise, except for very small objects \cite{borges2020schrodinger}. Of course measurements that involve further processing of plant parts, as often done in traditional morphological studies on herbarium specimens, are not possible from images.

Automated pathogen detection on collection material will also profit from the segmentation of plant organs from Herbarium sheet images, as many pathogens or symptoms of a plant disease only occur on specific organs. Studies on gall midges \cite{veenstra2012herbarium} have found herbarium specimens to be interesting study objects and would potentially profit from computer vision.

Manual annotation of herbarium specimens with bounding boxes as done for the training and test datasets in this study is a rather time-consuming process. Verification and correction of automatically annotated specimens is considerably faster, especially if the error rate is low. By iteratively incorporating expert-verified computer-generated data into new training datasets, the results can be further improved with reasonable efforts using Continual Learning \cite{parisi2019continual} .

\section*{Acknowledgments}

SY, MS and SD received funding from the DFG Project Mobilization of trait data from digital image files by deep learning approaches (grant 316452578). We gratefully acknowledge the support of NVIDIA Corporation with the donation of the TITAN Xp GPU to CW used for this research. Digitization of the Senckenberg specimens used in this study has taken place in the frame of the Global Plants Initiative.

\section*{Author Contributions}

\textbf{Sohaib Younis} is computer scientist at Senckenberg Biodiversity and Climate Research Center with focus on deep learning and image processing. Contributions: convolutional network modeling, image preprocessing, annotation of herbarium scans, organ detection, description of results and preparation of the manuscript.

\textbf{Marco Schmidt} is botanist at Senckenberg Biodiversity and Climate Research Center (SBIK-F) with a focus on African savannas and biodiversity informatics (eg online databases like African Plants - a photo guide and West African vegetation) and is working at Palmengarten’s scientific service, curating living collections and collection databases. Contributions: concept of study, annotation and verification of herbarium scans, preparation of the manuscript.

\textbf{Claus Weiland} is scientific programmer at SBIK-F’s Data \& Modelling Centre with main interests in large-scale machine learning, trait semantics and scientific data management. Contributions: Design of the GPU platform, data analysis and preparation of the manuscript.

\textbf{Stefan Dressler} is curator of the phanerogam collection of the Herbarium Senckenbergianum Frankfurt/M., which includes its digitization and curation of associated databases. Taxonomically he is working on Marcgraviaceae, Theaceae, Pentaphylacaceae and several Phyllanthaceous genera. Contribution: Herbarium Senckenbergianum collection, preparation of the manuscript.

\textbf{Bernhard Seeger} is professor of computer science systems at the Philipps University of Marburg. His research fields include high-performance database systems, parallel computation and real-time processing of high-throughput data with a focus on spatial biodiversity data. Contribution: Provision of support in machine learning and data processing.

\textbf{Thomas Hickler} is head of SBIK-F’s Data \& Modelling Centre and Professor for Biogeography at the Goethe University Frankfurt. He is particularly interested in interactions between climate and the terrestrial biosphere, including potential impacts of climate change on species, ecosystems and associated ecosystem services. Contribution: Preparation of the manuscript, comprehensive concept of study within biodiversity sciences.

\section*{Conflict of interest}

No potential conflict of interest was reported by the authors.

\bibliographystyle{plain}
\bibliography{ms}  


\end{document}